\documentclass[10pt, a4paper]{article}

\usepackage[final]{lrec-coling2024} % this is the new style
\usepackage{multibib}

\usepackage{times}  % DO NOT CHANGE THIS
\usepackage{helvet}  % DO NOT CHANGE THIS
\usepackage{courier}  % DO NOT CHANGE THIS
\usepackage{caption} % DO NOT CHANGE THIS AND DO NOT ADD ANY OPTIONS TO IT
\usepackage{subcaption}

\usepackage{algorithm}
\usepackage{algorithmic}
\usepackage{framed}
\usepackage{multirow}
\usepackage{booktabs}
\usepackage{xcolor}

\newcommand{\encoder}[1]{$E$}
\newcommand{\decoder}[1]{$D$}
\newcommand{\maskoutputv}[1]{$\hat{x}_v$}
\newcommand{\methodname}[1]{UDOP}
\newcommand{\rvlcdip}[1]{RVL-CDIP}
\newcommand{\publaynet}[1]{PubLayNet}
\newcommand{\tablebank}[1]{TableBank}
\newcommand{\docbank}[1]{DocBank}
\newcommand{\websrc}[1]{WebSRC}
\newcommand{\visualmrc}[1]{VisualMRC}
\newcommand{\iitcdip}[1]{IIT-CDIP}
\newcommand{\iitcdipfull}[1]{IIT-CDIP Test Collection 1.0}
\newcommand{\funsd}[1]{FUNSD}
\newcommand{\cord}[1]{CORD}
\newcommand{\duebenchmark}[1]{DUE-Benchmark}

\newcommand{\docvqa}[1]{DocVQA}
\newcommand{\infovqa}[1]{InfographicsVQA}
\newcommand{\kleister}[1]{Kleister Charity}
\newcommand{\pwc}[1]{PWC}
\newcommand{\deepform}[1]{DeepForm}
\newcommand{\wtq}[1]{WTQ}
\newcommand{\tabfact}[1]{TabFact}

\title{LayoutLLM: Large Language Model Instruction Tuning 
for Visually Rich Document Understanding}
\name{Masato Fujitake} 

\address{
FA Research, 
Fast Accounting Co., Ltd.
\\
fujitake@fastaccounting.co.jp
}

\abstract{
This paper proposes LayoutLLM, a more flexible document analysis method for understanding imaged documents.
Visually Rich Document Understanding tasks, such as document image classification and information extraction, have gained significant attention due to their importance.
Existing methods have been developed to enhance document comprehension by incorporating pre-training awareness of images, text, and layout structure.
However, these methods require fine-tuning for each task and dataset, and the models are expensive to train and operate.
To overcome this limitation, we propose a new LayoutLLM that integrates these with large-scale language models (LLMs).
By leveraging the strengths of existing research in document image understanding and LLMs' superior language understanding capabilities, the proposed model, fine-tuned with multimodal instruction datasets, performs an understanding of document images in a single model.
Our experiments demonstrate improvement over the baseline model in various document analysis tasks.
 \\ 
 \newline 
}

\begin{document}

\maketitleabstract

\section{Introduction} \label{sec:intro}
\begin{figure}
    \centering
    \begin{subfigure}{.20\textwidth}
        \includegraphics[width=\linewidth]{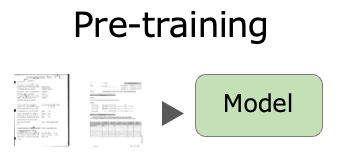}
%        \label{fig:key_idea_pre}
    \end{subfigure}\hfill 
    \\
    \vspace*{1.00\baselineskip}
    \begin{subfigure}{.50\textwidth}
        \includegraphics[width=0.9\linewidth]{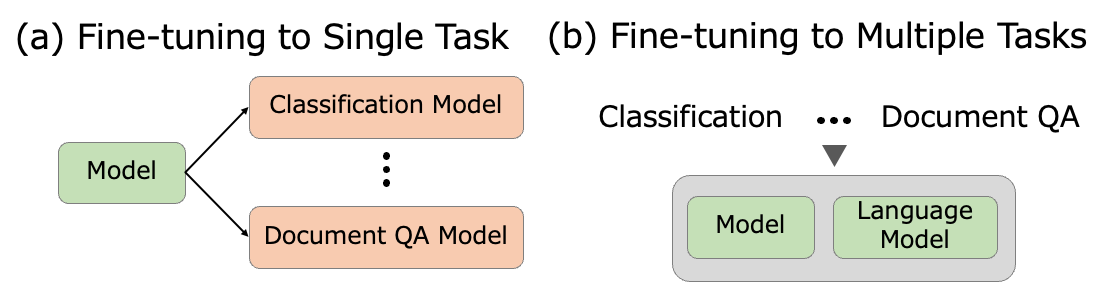}
%        \label{fig:key_idea_fine}
    \end{subfigure}\hfill
%    {
%    \small
    \caption{
    The overview of the existing and proposed method approaches.
The current method is fine-tuned for each task after pre-training, as shown in (a).
In contrast,  our method is fine-tuned via LLMs to handle multiple tasks, as shown in (b).
    }
%    }
    \label{fig:key_idea}
\vspace*{-1.00\baselineskip}
\end{figure}

Visual-rich Document Understanding (VrDU) focuses on analyzing document images, such as invoices, to extract and organize structured information automatically.
Different documents have different styles, formats, and contents, so unlike traditional textual information extraction tasks, VrDU relies on both textual and visual information.
Therefore, taking advantage of the multimodal nature of visually rich documents is essential.
To this end, previous works, such as LayoutLMs~\cite{huang2022layoutlmv3,xu2021layoutlmv2, xu2020layoutlm}, have proposed to acquire feature representations by jointly pre-training textual, visual, and layout information end-to-end in a single model, as shown in Figure~\ref{fig:key_idea}.
The process of fine-tuning is carried out on each task, as illustrated in Figure~\ref{fig:key_idea}(a).
However, this approach requires complex fine-tuning steps for each task and dataset, significantly increasing training and operational costs.

Large language models (LLMs) have gained a lot of attention due to their success in natural language processing tasks~\cite{brown2020gpt3}.
They acquire linguistic knowledge by predicting the continuation of input sentences through pre-training on large amounts of the corpus~\cite{radford2019language}.
Then, a model can perform various tasks, such as translation and summarization, by fine-tuning the knowledge with responses to the input text.
However, while they can perform various tasks through prompts, which are input instructions, they can only handle one-dimensional sequences of textual information.
They must be improved to handle text with significant two-dimensional structure, such as document images.

We propose a new approach, LayoutLLM, which tackles the limitations of conventional models by combining the advantages of VrDU models and LLMs.
As an encoder, it employs a model that excels in document layout understanding, while as a decoder, it uses LLMs that excel in language understanding.
The proposed single model can flexibly perform multiple tasks by fine-tuning to multiple VrDU tasks, as shown in Figure~\ref{fig:key_idea}(b).
We evaluated the proposed method on various benchmarks, such as document image classification, information extraction, and document visual question-answering.
Our experimental results confirm that LayoutLLM outperforms professionally tuned models in the VrDU task on several tasks and also improves performance on NLP tasks.

\section{Related Works} \label{sec:relatedwork}
\noindent
\textbf{VrDU.} 
Previously, language models~\cite{devlin2018bert} challenged document image analysis using only optical character recognition (OCR) text~\cite{fujitake2023diffusionstr, fujitake2023dtrocr}. 
However, a current approach integrates document images and OCR text to pre-train text, visual, and document layout, providing a more comprehensive understanding of documents.
LayoutLM~\cite{xu2020layoutlm} combines 2D location information, image embedding, and text for pre-training, like masking language modeling.
The improved models and pre-training techniques, such as LayoutLMv3~\cite{huang2022layoutlmv3}, have been proposed for higher accuracy~\cite{xu2021layoutlmv2, lee2023pix2struct}.
Text and document feature representations have been improved through multimodal encoders~\cite{gu2021unidoc}.
A method introduced modeling documents as a collection of bounding box tokens~\cite{garncarek2021lambert}.
OCR-free models generate text output directly from document images for optimization~\cite{kim2022donut, li2022dit, ye2023mplug}. 
UDOP~\cite{tang2023udop} is a method that combines multiple modes and tasks into a single model with image reconstruction. 
However, it only works for VrDU tasks and cannot handle NLP tasks. 
Therefore, this work proposes a flexible framework for multi-domain NLP and VrDU tasks by using pre-trained models as encoders and fine-tuning them with LLMs.

\noindent
\textbf{LLMs.} 
Large language models have been rapidly studied in recent years after the success of language models~\cite{devlin2018bert}.
BERT proposed a pre-training method, masked language modeling, which learns bidirectional text representations and then fine-tunes them to the target task.
GPT~\cite{radford2018gpt} also proposed a method for acquiring representations through next-word prediction pre-training.
The succeeding research found that by successfully inputting prompts, the model can be adapted to various tasks without fine-tuning~\cite{radford2019language}, and various large-scale language models have been proposed~\cite{brown2020gpt3, armengol2022multilingual}.
In this study, we used Alpaca~\citelanguageresource{alpaca}, based on the large language model Llama~\cite{touvron2023llama}, and fine-tuned with a dataset of 52K instructions and their responses.

\section{Method} \label{sec:method}

\begin{figure}[htp]
	\centering
	\includegraphics[width=1.00\columnwidth, keepaspectratio]{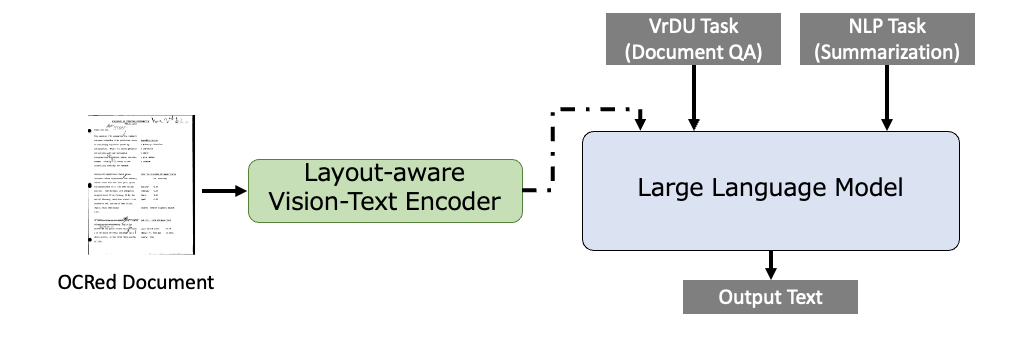}
	\caption{
Architectural overview of the proposed LayoutLLM.
It consists mainly of an encoder that encodes document images and a decoder that interprets tasks, and outputs.
	}
	\label{fig:overall_architecture}
     \vspace*{-0.50\baselineskip}
\end{figure}

Figure~\ref{fig:overall_architecture} shows an overview of the proposed method.
Our method, LayoutLLM, consists of pre-trained VrDU models and LLMs.
VrDU models, particularly LayoutLMv3~\cite{huang2022layoutlmv3}, handle visual and layout comprehension of documents.
The LLMs, especially Llama~\cite{touvron2023llama}, interpret and analyze the document's textual content and the task with the language understanding capabilities, and output the results.

The proposed method is fine-tuned using VrDU and NLP tasks.
As a training phase, the proposed method uses LayoutLMv3 as an encoder to process document images.
It takes an OCRed document image, which is a document image with OCR text by some methods~\cite{jaided2020EasyOCR, fujitake2023a3s}, as input and generates features.
Next, the model receives the document features and subsequent VrDU task instructions as prompts, and inputs them to Llama.
The response to the prompt is output in an autoregressive manner, and fine-tuning is performed using CrossEntropy in the same way as in standard LLMs training.
We follow Alpaca~\citelanguageresource{alpaca}, which fine-tunes Llama with Intruct-tuning for the NLP task.
Precisely, we fine-tune the model to the NLP task to respond appropriately, such as summarizing, based on the input of text prompts.
We do not use the VrDU model at this time and train in the same way as in Alpaca.

In the test phase, the proposed method uses the encoder to generate features for the VrDU task and input them to the LLM with the prompts, as in the training phase.
In the case of the NLP task, only text prompts are used to output responses.

\noindent
\textbf{Encoder.} 
OCRed textual and visual information from the document image is first encoded using the pre-trained LayoutLMv3 architecture.
It captures the layout structure and text of the document, and generates features corresponding to the document.
It is made into a 1D sequence with a maximum sequence length of 512 to be input to subsequent Llama.
The maximum sequence is determined by the LayoutLMv3 configuration.
In addition, one linear layer was used to correspond to the input dimension of Llama.
The feature is input to Llama.

\begin{table}[t]
\centering
\caption{Prompt format for VrDU tasks.}
\label{tab:prompt_example}
\small
%\begin{tabular}{p{1.3cm} p{13.5cm}} 
\begin{tabular}{p{7.4cm}} 
\toprule
{
\footnotesize
\textbf{Prompts}  }\\
\midrule
%Alpaca & 
{
%\small
\footnotesize
The previous information is about document images.
Below is an instruction that describes a task. 
Write a response that appropriately completes the request.

\#\#\# Instruction:
\{instruction\}

\#\#\# Response: 
}
\\

\bottomrule
\end{tabular}
% \end{adjustbox}
    \vspace*{-1.00\baselineskip}
\end{table}

\noindent
\textbf{Decoder.} 
The decoder uses Llama to receive input data and task instructions, and produces corresponding output.
It is an auto-regressive language model using the optimized Transformer architecture, and a 7B parameters model was used in this study.
Document features and VrDU task instruction prompts are used as input for the VrDU tasks.
More precisely, after tokenizing and encoding the task instruction prompts, the document features are first input to the LLM, followed by task instruction information.
The feature is input at the same stage as the features after embedding the natural language prompts. 
The NLP task fine-tuning method follows Alpaca completely.

\noindent
\textbf{VrDU Prompts.} 
Based on the document data, the task content is given to Llama by prompts to correspond to the target VrDU task.
The prompts are made as consistent as possible with the Alpaca format, and the format is shown in Table~\ref{tab:prompt_example}.
The ``\{instruction\}'' is a variable, the content of which is task-specific.
For document classification, we use ``Perform document classification. The classification labels are ...''
In the case of document information extraction, we use ``Perform document information extraction. The classification labels are... The output format is a set of extraction words and their labels, separated by commas. If multiple extraction targets exist, use \textbackslash n as a separator and split the outputs.''
For document questions,  we use ``Perform document question answering. The question is that ...''
Ground truth is created from each dataset to match the prompts' output.

\section{Experiments} \label{sec:experiments}

\begin{table*}[t]
\caption{
Performance comparison with state-of-the-arts on \funsd{}, \cord{}, and \rvlcdip{} datasets. 
Modality V, T, L denote vision, text and layout. 
}
\label{tab:main_results}
\centering
\resizebox{1.20\columnwidth}{!}{
\begin{tabular}{l c ccccc}
\toprule
Model & Modality & FUNSD & CORD & RVL-CDIP &Doc VQA \\
\midrule
BERT\textsubscript{large}~\cite{devlin2018bert}                 & T     & 65.6          & 90.3          & 89.9          & 67.5 \\
DiT\textsubscript{large}~\cite{li2022dit}                       & V     & $-$           & $-$           & 92.7          & $-$  \\
Donut~\cite{kim2022donut}                                       & V     & $-$           & 91.6          & 95.3          & 72.1 \\
mPLUG-DocOwl~\cite{ye2023mplug}                                 & V     & $-$           & $-$           & $-$           & 62.2 \\
%BROS\textsubscript{large}~\cite{hong2022bros}                   & T+L   & 84.5          & 97.4          & $-$           & $-$  \\
StructuralLM\textsubscript{large}~\cite{li2021structurallm}     & T+L   & 85.1          & $-$           & 96.2          & 83.9 \\
%LiLT\cite{wang2022lilt}                                         & T+L   & 88.4          & 96.1          & 95.7          & $-$  \\
%FormNet~\cite{lee2022formnet}                                   & T+L   & 84.7          & 97.3          & $-$           & $-$  \\
LayoutLM\textsubscript{large}~\cite{xu2020layoutlm}             & T+L   & 77.9          & $-$           & 91.9          & $-$  \\
%SelfDoc~\cite{li2021selfdoc}                                    & V+T+L & 83.4          & $-$           & 92.8          & $-$  \\
UniDoc~\cite{gu2021unidoc}                                      & V+T+L & 87.9          & 96.9          & 95.1          & $-$  \\
LAMBERT~\cite{garncarek2021lambert}                             & T+L   & $-$           & 96.1          & $-$           & $-$  \\
%DocFormer\textsubscript{large}~\cite{appalaraju2021docformer}   & V+T+L & 84.6          & 97.0          & 95.5          & $-$  \\
TILT\textsubscript{large}~\cite{powalski2021going}              & V+T+L & $-$           & 96.3          & 95.5          & \textbf{87.1} \\
LayoutLMv2\textsubscript{large}~\cite{xu2021layoutlmv2}         & V+T+L & 84.2          & 96.0          & 95.6          & 78.8 \\
LayoutLMv3\textsubscript{large}~\cite{huang2022layoutlmv3}      & V+T+L & 92.1          & 97.5          & 95.9          & 83.4 \\
UDOP~\cite{tang2023udop}                                        & V+T+L & 91.6          & 97.6          & 96.0          & 84.7 \\
%LayoutGCN~\cite{shi2023layoutgcn}                               & V+L   & 82.1          & 95.7          & 89.8          & $-$  \\
\midrule
LayoutLLM (Ours)                                                & V+T+L & \textbf{95.3} & \textbf{98.6} & \textbf{98.8} & 86.9 \\
\bottomrule
\end{tabular}
}
    \vspace*{-1.00\baselineskip}
%\end{table}
\end{table*}

\subsection{Dataset and Evaluation}
LayoutLLM's performance was evaluated through experiments such as form understanding, receipt recognition, and document classification tasks.
Unless stated otherwise, OCR text and bounding boxes are extracted by EacyOCR~\cite{jaided2020EasyOCR}.

\noindent
\textbf{Document Classification.} 
Document classification predicts the category of each document image.
RVL-CDIP~\citelanguageresource{harley2015evaluation} is used as the target dataset.
This dataset comprises 320K/40K/40K training/validation/test images in 16 categories.
Classification accuracies for the 16 categories are used to measure model performance.

\noindent
\textbf{Document Information Extraction.} 
To extract information from documents, a model must predict the label for each semantic entity. 
We use the FUNSD~\citelanguageresource{jaume2019funsd} and CORD~\citelanguageresource{park2019cord} datasets.
FUNSD uses 149/50 noisy document images during training and testing.
Each semantic entity includes a word list, label, and bounding box. 
The evaluation measure used is an entity-level F1 score for predicting a question, answer, header, or other.
We use the OCR text and bounding boxes provided by the dataset.

The CORD dataset is a benchmark for receipt comprehension, with 626/247 receipts for training/testing, respectively. 
A model must recognize a list of text lines.
The receipts are labeled with 30 entities grouped into four categories: company, date, address, and total.
The metric is F1, and the task format is the same as FUNSD.

\noindent
\textbf{Document Visual Question Answering.} 
We use a document understanding benchmark, DocVQA~\citelanguageresource{mathew2021docvqa}.
It consists of 50,000 questions defined on over 12,000 pages of documents. 
The dataset is organized into training, validation, and test sets, with a ratio of about 8:1:1.
It contains an OCRed image page, questions, and answers. 
The task is evaluated using an ANLS, an edit distance-based metric measuring average normalized Levenshtein similarity.

\subsection{Implementation Details}
Our model used a pre-trained LayoutLMv3 large encoder and a pre-trained Llama-7B decoder with their official weights. 
We used VrDU task datasets and NLP task datasets to fine-tune the model. 
We created the VrDU task dataset using the dataset above and the prompt described in the proposed method section. 
For the NLP task dataset, we used the Alpaca dataset~\citelanguageresource{alpaca}.
Mini-batches were created during training, separating the VrDU and NLP tasks. 
It was because the NLP task doesn't require data input to the encoder. 
Mixing the two tasks prevented back-propagation.
Encoder outputs and prompt inputs were consistently supplied in the same order for training and inference.
The model is optimized on eight A100 GPUs with a batch size of 16.
We follow Alpaca's learning process basically, using AdamW Optimizer~\cite{loshchilov2018adamw}, with a learning rate of 1e-5 and 20 epochs.
Cosine learning rate scheduling was used, with a warmup ratio of 0.05 and weight decay of 0.01.

\subsection{Main Results}
Table~\ref{tab:main_results} shows the performance of each dataset with several state-of-the-art methods.

\noindent
\textbf{Document Classification.} 
Our approach results in new state-of-the-art accuracy, surpassing StructuralLM's previous record by 2.6 points without task-specific and special fine-tuning after pre-training. 
The improvement in accuracy can be attributed to two factors: improved linguistic ability using LLMs and learning multiple tasks simultaneously. 
The conventional method only considers linguistic context obtained during pre-training and fine-tuning with documents. In contrast, the proposed method takes advantage of a language model specialized for linguistic information, making it easier to classify documents based on their content.

\noindent
\textbf{Document Information Extraction.} 
The proposed method has achieved outstanding results on both datasets, with scores of 95.3\% and 98.6\% on FUNSD and CORD, respectively. 
These scores represent a significant improvement of 3.2 points over LayoutLMv3's best accuracy for the FUNSD dataset and 1.0 points over UDOP's best accuracy for the CORD dataset. 
UDOP and our method are both a single model for VrDU tasks.
However, the proposed method uses a language model, extracting more accurate information from documents.

\noindent
\textbf{Document Visual Question Answering.} 
Our method achieved an accuracy of 86.9\%, which is comparable to the highest accuracy achieved in previous works.
The 3.5 points improved over the baseline show its effectiveness, as it uses a language model as a decoder, which enhances language comprehension in Q\&A sessions.

\subsection{Detailed Analysis}
We evaluated the method in detail using CORD and RVL-CDIP datasets.

\begin{table}[!t]
  \caption{
  Component analysis.
}
\label{tab:ablation_decoder_analysis}
  \centering
\resizebox{1.0\columnwidth}{!}{%0
  \begin{tabular}{lcc}
    \toprule
    Method &RVL-CDIP &CORD  \\
    \midrule
    Encoder-only (LayoutLMv3)                    & 95.9 & 97.5  \\
    $+$ Decoder (Llama) with Each VrDU Task           & 97.5 & 97.9  \\
    $+$ Decoder (Llama) with Multi VrDU Tasks            & 98.1 & 98.1  \\
    $+$ Decoder (Llama) with Multi VrDU \& NLP Tasks    & \textbf{98.8} & \textbf{98.6}  \\
  \bottomrule
\end{tabular}
}
    \vspace*{-1.00\baselineskip}
\end{table}

\noindent
\textbf{Component Analysis.} 
We performed a stepwise validation to see whether the proposed method works.
Table~\ref{tab:ablation_decoder_analysis} shows the results of fine-tuning to each task using only the encoder, fine-tuning to each VrDU task only using the decoder, fine-tuning to multiple VrDU tasks simultaneously, and finally fine-tuning with the NLP task.
We confirm that incorporating the language model, VrDU multi-task training, and NLP multi-task training are all crucial.

\begin{table}[bt]
\centering
\caption{
	Impact of encoder.
}
\label{tab:ablation_encoder_analysis}
\resizebox{0.7\columnwidth}{!}{%0
\begin{tabular}{c|c|cc}
\toprule
Encoder    & Modal   & RVL-CDIP & CORD  \\ \midrule
LayoutLMv3 & V+T+L & \textbf{98.8} & \textbf{98.6} \\
UniDoc  & V+T+L & 98.1 & 97.9 \\
DiT        & V     & 94.3 & 89.6 \\

\bottomrule
\end{tabular}
}
    \vspace*{-1.00\baselineskip}
\end{table}

\noindent
\textbf{Effects of Encoder Component.} 
We evaluated the impact of the encoder component.
Our approach is a flexible framework, and the encoder can be replaced.
Table~\ref{tab:ablation_encoder_analysis} shows the results of replacing the encoder with other methods.
The encoder's feature output length is set to 512.
It shows the successful performance of various methods and modalities.
This suggests that more robust methods in the future can be incorporated flexibly.

\begin{table}[htbp]
\centering
\caption{Affect on the NLP task performance.}
\label{tab:comparison_openllm}
\resizebox{1.0\columnwidth}{!}{%
\begin{tabular}{c|r|r|r|r|r}
\toprule
Model          & Average & ARC      & HellaSwag   & MMLU    & TruthfulQA  \\ \midrule
Alpaca         & 52.02   & 52.05    & 77.00       & 41.45   & \textbf{37.60}       \\ \
Proposed model & \textbf{53.06}   & \textbf{52.12}    & \textbf{79.32}       & \textbf{44.31}   & 36.49       \\ \bottomrule
\end{tabular}
}
    \vspace*{-1.00\baselineskip}
\end{table}

\noindent
\textbf{Impact on the NLP Tasks.}
We used a generic LLM as a decoder and fine-tuned it with a regular NLP task and the VrDU task.
We investigated how learning the NLP and VrDU tasks together affects the NLP task.
We used various benchmarks that have been used in recent years to evaluate LLMs. We used the following benchmarks: ARC~\cite{chollet2019measure} for multiple-choice, HellaSwag~\cite{zellers2019hellaswag} for sentence completion, MMLU~\cite{hendrycks2020measuring} for multidomain knowledge understanding, and Truthful TruthfulQA~\cite{lin2021truthfulqa}, which measures the accuracy of answers to questions.
The results of the Alpaca model and the proposed method are presented in Table~\ref{tab:comparison_openllm}.
Although the VrDU task appeared to negatively impact the NLP task because it is in a different domain, the average score increased.
In particular, it improved by 2.86 points in language comprehension with MMLU.
Further research is needed to explore which NLP tasks, such as summarization, are associated with higher scores with VrDU.

\section{Conclusion}\label{sec:conclusion}
This study has presented a document analysis framework capable of performing multiple tasks.
The proposed approach, LayoutLLM, combines a VrDU encoder to capture document images and a decoder, LLM, to receive task instructions and process them accordingly.
It allows us to efficiently understand document images by capturing visual and textual context.
Experimental results show that our method significantly improves the performance of various VrDU tasks.
Furthermore, unlike previous studies, it can exploit LLMs' pure NLP task processing capability, not only for VrDU tasks.

\newpage

\nocite{*}
\section{Bibliographical References}\label{sec:reference}
\bibliographystyle{lrec-coling2024-natbib}
\bibliography{reference}

\section{Language Resource References}
\label{lr:ref}
\bibliographystylelanguageresource{lrec-coling2024-natbib}
\bibliographylanguageresource{languageresource}

\end{document}